\documentclass{acm_proc_article-sp}

\begin{document}

\title{Beyond Spatial Pyramid Matching: Space-time Extended Descriptor for Action Recognition}

\numberofauthors{1} 
\author{
\alignauthor
Zhenzhong Lan, Alexander G. Hauptmann \\
 Carnegie Mellon University\\
{\tt\small  \{lanzhzh, alex\}@cs.cmu.edu}
}

\date{20 April 2013}

\maketitle
\begin{abstract}
We address the problem of generating video features for action recognition. The spatial pyramid and its variants have been very popular feature models due to their success in balancing spatial location encoding and spatial invariance. Although it seems straightforward to extend spatial pyramid to the temporal domain (spatio-temporal pyramid), the large spatio-temporal diversity of unconstrained videos and the resulting significantly higher dimensional representations make it less appealing. This paper introduces the space-time extended descriptor, a simple but efficient alternative way to include the spatio-temporal location into the video features. Instead of only coding motion information and leaving the spatio-temporal location to be represented at the pooling stage, location information is used as part of the encoding step. This method is a much more effective and efficient location encoding method as compared to the fixed grid model because it avoids the danger of over committing to artificial boundaries and its dimension is relatively low. Experimental results on several benchmark datasets show that, despite its simplicity, this method achieves comparable or better results than spatio-temporal pyramid.

\end{abstract}

% A category with the (minimum) three required fields


 % NOT required for Proceedings

\section{Introduction}

\begin{figure}
\centering
\begin{tabular}{cc}
\includegraphics[height = 1.8cm, width=4.0cm]{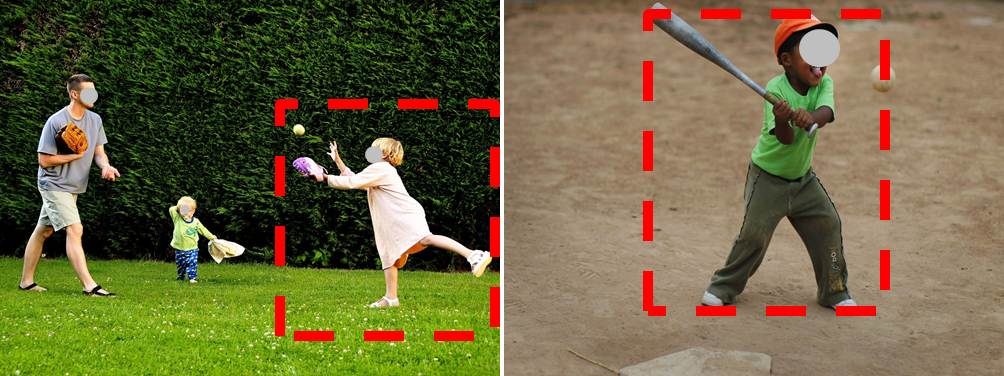}&
\includegraphics[height = 1.8cm, width=4.0cm]{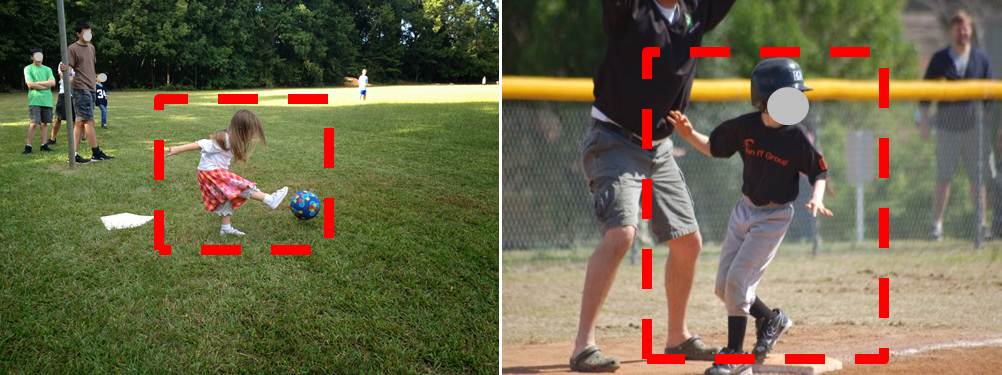} 
\end{tabular}
\caption{ ``Catch'' and ``hit'' are likely to be distinguished
by upper-bodies, especially hands while
 ``kick'' and ``run'' are more easily distinguished by legs.}
\label{fig:examples}
\end{figure}

\begin{figure}
\centering
\begin{tabular}{cc}
\includegraphics[height = 1.8cm, width=4.0cm]{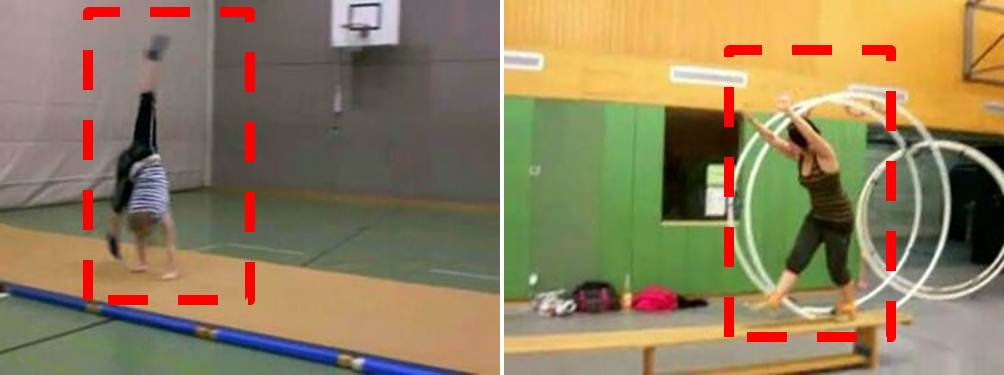}&
\includegraphics[height = 1.8cm, width=4.0cm]{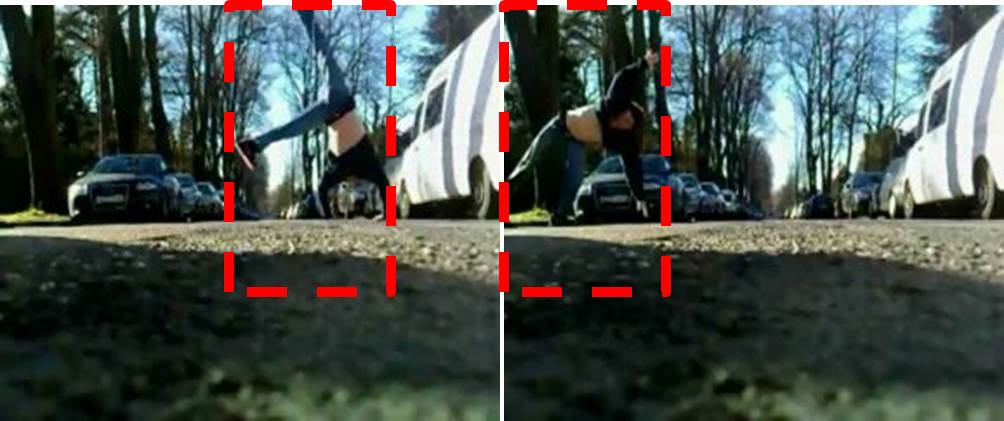} \\
Inter-videos & Intra-video
\end{tabular}
\caption{In different videos, actions localization can be
subject to variation due to camera viewpoint change. But,
even within a single video sequence, the action area can
change among frames.}
\label{fig:e2}
\end{figure}

With the constant expansion of on-line video collections, content-based video retrieval \cite{jiang2007towards,amir2003ibm,snoek2008concept, lan2013cmu}  has become an important problem in multimedia retrieval. It is a difficult task since on-line videos are subject to large visual diversity. Robust to such variability, the Bag-of-Features (BoF) [23] model has been used as the main paradigm for representing videos. A BoF can be summarized as an encoding step and a pooling step \cite{boureau2010learning}. Traditional pooling discards the local feature position information in the video space. However, this spatio-temporal information has been proven to be a discriminative cue \cite{laptev2008learning}. Indeed, discriminative motion information is not equally distributed in the video space as shown in Figure \ref{fig:examples}. To benefit from this information, spatial pyramids \cite{lazebnik2006beyond, laptev2008learning} divides a video using fixed grids and pools the features locally into each grid cell. Spatial pyramids can be easily extend to spatio-temporal pyramid (STP) to encode the order of actions or events, another important cue for video representation. For example, to perform an action called ``sitting down'', we need to gradually bend our knees and lower our body, while ``standing up'' is the reverse. Despite its usefulness, STP is not as effective as its spatial sibling  on still images due to the fact that actions in unconstrained videos will have much more dramatic spatio-temporal variance than still images. For example, as illustrated in Figure 2, we see the person doing a cartwheel moves through out the frame whereas a still image would center the person in the frame. In this case, STP, which commits too much to the artificial boundaries may lead to a performance drop. Another major criticism of spatial pyramid is that it generates features with dimensions that are orders of magnitude higher than the original spatio-temporal invariant representations and hence make it computationally expensive to process. A more effective and efficient space-time encoding method is therefore critical for video representations in retrieving videos from large data collections. 

In this work, we propose to take advantage of the spatio-temporal discriminative information with an emphasis on retaining the spatio-temporal robustness and controlling dimension explosion. Beyond standard spatial pooling which uses fixed segmentation grids, our proposed space-time extended descriptor (STED) augments the feature descriptor with spatio-temporal location information.  By simultaneously encoding motion and location information, we remove the necessity of a complicated pooling stage and the danger of committing to artificial boundaries. Experimental results on several benchmark datasets show the advantages of STED over STP. 

In the remainder of this paper, we start by providing more background information about spatial pyramids and its variants. We then compare STP and STED in detail. After that, an evaluation of our method is performed. Further discussions including potential improvements are given at the end.

\section{Related Work}
Video retrieval research is conducted in a diverse setting where emphasis includes low-level and high-level feature design \cite{snoek2008concept, jiang2007towards}, multi-modality fusion \cite{amir2003ibm, lan2013cmu} and multi-modality retrieval models \cite{jiang2014zero}. Here we focus our review on low-level features design and encoding which is related to our
latter experimental comparison. There has been a large amount of work in trying to build representations that keep spatial information of image patterns. Among them, spatial pyramid matching \cite{lazebnik2006beyond} is the most popular one. However, building spatial pyramids requires dimensions that are orders of magnitude higher than the original spatial invariant representations and hence make it less suitable for high dimensional encoding methods such as Fisher vector \cite{perronnin2010improving} and VLAD \cite{arandjelovic2013all}. Spatial Fisher vector \cite{krapac2011modeling} and spatial augmentation \cite{mccann2013spatially, sanchez2012modeling} provide more compact representations to encode spatial information and show similar performance as spatial pyramid methods. Few approaches consider encoding global temporal information into video representations. Oneata et al. \cite{oneata2013action} show that better action recognition performance can be achieved by dividing videos into two parts and encoding each one separately. Codella et al. \cite{codella2012video} try to use temporal pyramid for event detection. They use $n$ temporal segments, where $n$ incrementally increases from 1 to 10.

\section{Space-time Encoding Methods}

In the following, we reformulate the spatial pyramid model and then extend this formulation to describe the STP and TED. 

Let $D = \{d_1(\phi_1,x_1,y_1), d_2(\phi_2,x_2,y_2),..., d_M(\phi_M,x_M,y_M)\}$ be a set of local feature descriptors extracted from a video. Each $d_i$ contains the appearance/motion description $\phi_i$, which is the typically local image descriptor,  and the normalized pixel location $(x_i, y_i)$ at which feature $i$ is centered. We denote function $g: \phi_i \rightarrow \mathbb{R}^K$ as a local feature encoding scheme such as sparse-coding or locality encoding. Note that here $g$ solely relies on the appearance/motion portion of the descriptors.  We further denote by $G = \{G_1, ..., G_n\}$ a set of grid cells. Each $G_j$ is a binary matrix indicating which pixel is active, $G_j \in {0, 1}^{(s_x \times s_y)}$ , $(s_x, s_y)$ being the image size. Based on those definitions, we express the average spatial pooling operation as 
\begin{align}
X_j = \frac{1}{n} \sum_{i=1}^M G^j_{(x_i, y_i)} \times g(\phi_i).
\label{f1}
\end{align}

\subsection{Spatio-Temporal Pyramid (STP)}

For temporal expansion, we simply add the temporal location into formula \ref{f1}. That is 
\begin{align}
X_j = \frac{1}{n} \sum_{i=1}^M G^j_{(x_i, y_i, t_i)} \times g(\phi_i). 
\label{f2}
\end{align}

Note that by having one more location dimension, the resulting number of grid cells in $G$, hence number of dimension in feature $X$, will be orders of magnitude higher than the original spatial pyramid method. This may not be computationaly affordable for high dimensional representation such as Fisher Vector. For example, if we want to represent an Improved Dense Trajectory (IDT) \cite{wang2011action} feature (dimension of $\phi_i$ is 426) using a Fisher vector representation with 256 Gaussian mixture models, then a $k \times k \times l$ with $k \in (1, 2)$ and $l \in (1,3)$  STP can result in a representation about 4.4 million dimensions.

\subsection{Space-Time Extended Descriptor (STED)}
\label{sted}
For STED, instead of using space-time information to locate the grid cells $G$, we use it to encode the feature. Formally, we have, 
\begin{align}
X = \frac{1}{n} \sum_{i=1}^Mg(\phi_i,x_i,y_i,t_i). 
\label{f3}
\end{align}

One advantage of STED is that we avoid having to commit to artificial grid boundaries to define the spatial pooling regions, which can lead to very divergent representations for similar actions happenning in different space-time location. Another advantage is that the dimension of STED is only slightly higher than the original space-time invariant representation and is much lower than STP. Again taking the Dense Trajectory + Fisher vector setting for example, for each video, STED generates a feature with only about 0.21 million dimensions, which is about 20 times lower than the STP representation.

\section{Experiments}

\begin{table*}
\centering
\begin{tabular}{|c| c| c |c |c | c|c |c |c|}
\hline
  & \multicolumn{2}{|c|}{UCF50}  & \multicolumn{2}{|c|}{HMDB51} & \multicolumn{2}{|c|}{Hollywood2} & \multicolumn{2}{|c|}{Olympic} \\
  & \multicolumn{2}{|c|}{(mAcc. $\%$)} & \multicolumn{2}{|c|}{(mAcc. $\%$)} & \multicolumn{2}{|c|}{(mAP $\%$)} & \multicolumn{2}{|c|}{(mAP $\%$)}\\ \hline
  $l$ & Single& Pyramid & Single & Pyramid & Single& Pyramid &Single& Pyramid\\
  \hline
 1 &\textbf{92.7} &  & 59.6& & 65.8 & & \textbf{89.8} & \\
 2 &92.3& 92.6  & 61.0& 61.3 & 66.3 & \textbf{67.4} & 87.8 & 89.3\\
 4 &91.6  & 92.3 & 60.5& \textbf{61.6} & 65.2 & 67.0 & 85.2 & 87.6\\
 8 &90.4  & 92.1 & 58.1& \textbf{61.6} & 62.2 & 65.7 & 83.8 & 83.8\\
\hline
\end{tabular}
\caption{\label{tab:temporal pyramids}Comparison of different temporal pyramid levels for STP.}
\end{table*}

\subsection{Experimental Setting}

IDT with Fisher vector encoding \cite{wang2013action} represents a current state-of-the-art for most real-world action recognition datasets. Therefore, we use it to evaluate our method. Note that although we use Fisher vector, our method can be applied to any quantization and pooling method such as VLAD \cite{arandjelovic2013all}. Our baseline method uses the  same settings as in \cite{wang2013action}. These settings include the IDT feature extraction, Fisher vector representation and a linear SVM classifier.  

IDT features are extracted using 15 frame tracking, camera motion stabilization with human masking and RootSIFT normalization and described by Trajectory, HOG, HOF and MBH descriptors. We use PCA to reduce the dimensionality of these descriptors by a factor of two. For Fisher vector representation,  we map the raw feature descriptors into a  Gaussian Mixture Model with 256 Gaussians trained from a set of randomly sampled 256000 data points. Power and L2 normalization are also used before concatenating different types of descriptors into a video based representation. For classification, we use a linear SVM classifier with a fixed C=100 as recommended by \cite{wang2013action} and the one-versus-all approach is used for multi-class classification scenario. 

For STP, we use $k \times k \times l$, $k$ is 1 and 2 and $l \in (1,2,4,8)$ grid cells.  For STED, we attached the normalized 3 dimensional location to each descriptor as described in section \ref{sted}.

\subsection{Datasets}

We use four action retrieval or classification datasets, UCF50, HMDB51, Hollywood2 and Olympic Sports, for evaluation. 
These datasets, which mainly involves actions, are selected because they are well known real-world datasets.
The UCF50 dataset \cite{reddy2013recognizing} has 50 action classes spanning over 6618 YouTube video clips that can be split into 25 groups. Video clips in the same group are generally very similar in background. Leave-one-group-out cross-validation as recommended by \cite{reddy2013recognizing} is used and mean accuracy (mAcc) over all classes and all groups is reported.
The HMDB51 dataset \cite{kuehne2011hmdb} has 51 action classes and 6766 video clips extracted from digitized movies and YouTube. \cite{kuehne2011hmdb} provides both original videos and stabilized ones. We only use original videos in this paper and standard splits with mAcc are used to evaluate the performance. 
The Hollywood2 dataset \cite{marszalek2009actions} contains 12 action classes and 1707 video clips that are collected from 69 different Hollywood movies. We use the standard splits with training and test videos provided by \cite{marszalek2009actions}.   Mean average precision (mAP) is used to evaluate this dataset because multiple labels can be assigned to one video clip. 
The Olympic Sports dataset \cite{niebles2010modeling} consists of 16 athletes
practicing sports, represented by a total of 783 video clips. We use the standard split with 649 training clips and 134 test clips and report mAP as in \cite{niebles2010modeling} for comparison purposes. Note that under this setting, on average, each class only has 8 testing samples, which may be too few to give a concrete measurement of the model. 

\subsection{Experimental Results}

\subsubsection{Spatial-Temporal Pyramids (STP)}
In Table \ref{tab:temporal pyramids}, we compare the performance of the STP at different temporal pyramid levels $l$. $l=1$ corresponds to results only using a $1 \time 1 $ and a $2 \times 2$ spatial pyramid pooling. From Table  \ref{tab:temporal pyramids}, we can see that, due to the large spatio-temporal diversity of unconstrained videos, the usefulness of STP is inconclusive. For some datasets such as HMDB51 and Hollywood2, it provides significant performance improvement while for other datasets such as UCF50 and Olympic Sports, it hurts the performance.   Further division (from level 2 to 4 or 8) almost always results in worse performance with the exception of the pyramids in the HMDB51 dataset. These results show that a straightforward extension of spatial pyramids may hurt the performance.

\subsubsection{Space-time Extension Descriptor (STED)}
From Table \ref{tab:temporal augmentation}, we see that, unlike STP, STED consistently improves the performance for all datasets. It is worth mentioning that both HMDB51 and Hollywood2 are very challenge datasets, more than $3\%$ absolute improvement over space-time invariant representation is quite a notable gain. For Olympics dataset, because on average, each class only contains 8 testing examples, the improvement may not be statistically meaningful.

\begin{table}
\centering
\begin{tabular}{|c| c|c |c |c|}
\hline
  & UCF50  & HMDB51 & Hollywood2 & Olympic \\
 STED & (mAcc. $\%$)& (mAcc. $\%$) & (mAP $\%$) & (mAP $\%$)\\ \hline
 w/o &91.5 & 59.0 & 64.6 & 89.5  \\
 w & \textbf{93.0} &  \textbf{62.1} & \textbf{67.0} & \textbf{89.8} \\
 \hline
\end{tabular}
\caption{\label{tab:temporal augmentation}Performance of STED.}
\end{table}

\begin{table*}
\centering
\footnotesize
\begin{tabular}{|l c |l c |l c|l c|}
\hline
    \multicolumn{2}{|c|}{HMDB51 (MAcc. $\%$)} & \multicolumn{2}{|c|}{Hollywood2 (MAP $\%$)}  & \multicolumn{2}{|c|}{UCF50 (MAcc. $\%$)} & \multicolumn{2}{|c|}{Olympics Sports (MAP $\%$)}\\ \hline
    
Oneata \textit{et al.} \cite{oneata2013action}  &54.8  & Sapienz \textit{et al.} \cite{sapienza2014feature}  & 59.6   & Shi \textit{et al.} \cite{shi2013sampling}  & 83.3  & Jain \textit{et al.} \cite{jain2013better}  & 83.2\\
Wang \& Schmid \cite{wang2013action}   &57.2  &Jain \textit{et al.} \cite{jain2013better}    & 62.5   & Ciptadi \textit{et al.} \cite{ciptadi2014movement}  & 90.0 &   Adrien \textit{et al.} \cite{gaidon2014activity}  & 85.5 \\
Simonyan \textit{et al.} \cite{simonyan2014two}  & 57.9 &  Oneata \textit{et al.} \cite{oneata2013action} &  63.3    & Oneata \textit{et al.} \cite{oneata2013action} &90.0 & Oneata \textit{et al.} \cite{oneata2013action}  & 89.0\\
Peng \textit{et al.} \cite{peng2014bag}  & 61.1 & Wang \& Schmid \cite{wang2013action}  &64.3  & Wang \& Schmid \cite{wang2013action}  & 91.2 &  Wang \& Schmid \cite{wang2013action}  & \textbf{91.1}\\
\hline 
STP ( $l=2$ ) & 61.3 & STP( $l=2$ ) &  \textbf{67.4} & STP ( $l=2$ ) & 92.6 & STP( $l=2$ ) & 89.3 \\ 
STED & \textbf{62.1} & STED &  67.0 & STED & \textbf{93.0} & STED & 89.8\\ 
\hline
\end{tabular}
\caption{\label{tab:state-of-art}Comparison of our results to the state-of-the-arts.}
\end{table*}

\subsubsection{Comparing with the State-of-the-Art}
In Table \ref{tab:state-of-art},  we compare STP and STED, along with other recently published approaches. From Table \ref{tab:temporal pyramids}, we see that a temporal level of 2 gives stable results for STP, therefore we set the temporal level as 2. From Table \ref{tab:state-of-art}, we can see that STED achieves similar or better results in all four datasets. Considering that STED requires much lower dimensional encoding hence much less memory and computational cost than STP, it is a better space-time encoding method to use. 

Note that although we list several recent approaches here for comparison purposes, they do use different features or settings. Shi et al. \cite{shi2013sampling} use random sampled feature points and HOG, HOF, HOG3D and MBH descriptors. Jain et al. \cite{jain2013better}'s approach incorporates  a new motion descriptor. Oneata et al.  \cite{oneata2013action} focus more on testing Spatial Fisher vector for multiple action and event tasks. Ciptadi et al. \cite{ciptadi2014movement} presented a novel action representation based on encoding the global temporal movement of an action. The most comparable one is Wang et al. \cite{wang2013action}, from which we build our approaches and which serves as our baseline. Again, STED achieves comparable or better results than these state-of-art methods.

\section{Discussion}

In this paper, we propose STED, a simple extension of local descriptors with spatio-temporal information. Despite it simplicity, STED is a much more effective and efficient way to to encode space-time location than previous methods. By simultaneously coding appearance, motion and location, STED avoids the danger of committing to artificial grid boundaries that define the spatial-temporal pooling regions and hence is better in dealing with unconstrained video that have large spatial-temporal motion diversity. We compare STED with STP, a straightforward way to extend spatial pyramids, and show that STED generates representations with much lower dimension while achieving similar or better results. Further improvements include determining optimal weighting for appearance/motion and location. Also, implementing STED with additional location encoding methods.

\bibliographystyle{abbrv}
\bibliography{sample}

%\balancecolumns 

\end{document}